\documentclass{article}

\usepackage{arxiv}

\usepackage[utf8]{inputenc} % allow utf-8 input
\usepackage[T1]{fontenc}    % use 8-bit T1 fonts
\usepackage{hyperref}       % hyperlinks
\usepackage{url}            % simple URL typesetting
\usepackage{booktabs}       % professional-quality tables
\usepackage{amsfonts}       % blackboard math symbols
\usepackage{nicefrac}       % compact symbols for 1/2, etc.
\usepackage{microtype}      % microtypography
\usepackage{lipsum}
\usepackage{graphicx}
\graphicspath{ {./images/} }
\usepackage{amsmath}
\usepackage{amssymb}
\usepackage[inline]{enumitem}
\usepackage{xcolor}
\usepackage{makecell,multirow}
\usepackage{algorithm}
\usepackage{algorithmic}
\usepackage{rotating}

\newcommand\Mu{\mathcal{M}}
\newcommand\D{\mathcal{D}}
\newcommand{\STAB}[1]{\begin{tabular}{@{}c@{}}#1\end{tabular}}

\title{Dominant Set-based Active Learning \\for Text Classification and its Application \\to Online Social Media}

\author{
 Toktam A. Oghaz\textsuperscript{\rm 1}, Ivan Garibay\textsuperscript{\rm 1,}\textsuperscript{\rm 2} \\
}

\begin{document}
\maketitle
\begin{abstract}
Recent advances in natural language processing (NLP) in online social media are evidently owed to large-scale datasets. However, labeling, storing, and processing a large number of textual data points, e.g., tweets, has remained challenging. On top of that, in applications such as hate speech detection, labeling a sufficiently large dataset containing offensive content can be mentally and emotionally taxing for human annotators. Thus, NLP methods that can make the best use of significantly less labeled data points are of great interest. In this paper, we present a novel pool-based active learning method that can be used for the training of large unlabeled corpus with minimum annotation cost. For that, we propose to find the dominant sets of local clusters in the feature space. These sets represent maximally cohesive structures in the data. Then, the samples that do not belong to any of the dominant sets are selected to be used to train the model, as they represent the boundaries of the local clusters and are more challenging to classify. Our proposed method does not have any parameters to be tuned, making it dataset-independent, and it can approximately achieve the same classification accuracy as full training data, with significantly fewer data points. Additionally, our method achieves a higher performance in comparison to the state-of-the-art active learning strategies. Furthermore, our proposed algorithm is able to incorporate conventional active learning scores, such as uncertainty-based scores, into its selection criteria. We show the effectiveness of our method on different datasets and using different neural network architectures.  
\end{abstract}

% keywords can be removed

\section{Introduction}
\noindent 
The growing volume of user-generated content on online social media, and the necessity of moderating these platforms to protect the users from toxicity, cyberbullying, and trolling has led to conducting substantial efforts and analysis to find the indicator attributes \cite{mathew2019spread,davidson2017automated}, preparing manually labeled datasets for toxicity type classification and detection \cite{founta2018large,de2018hate,wulczyn2017wikipedia}, the development of machine learning models and systems for toxic language detection \cite{sahlgren2018learning,mozafari2020hate}, and automatic summarization techniques that help to grasp the main ideas from online social media data \cite{tarnpradab2021improving,oghaz2020probabilistic}. Although collecting a large unlabeled textual corpus is cheap, manually annotating a sufficiently large dataset for the training of a machine learning model is expensive. Besides the annotation cost, datasets for toxicity classification may suffer from annotation mistakes and difficulties, such as enduring bias toward some attributes and races, and containing significantly imbalanced training samples \cite{mozafari2020hate,chuang2021mitigating}. Additionally, the annotators of such datasets may undergo extra discomfort as they get exposed to offensive and abusive content such as hate speech. 

Active Learning (AL) techniques can mitigate the issues associated with manual labeling and improve automatic detection and classification when labeled training data is sparse \cite{dor2020active}. The goal of active learning is to \begin{enumerate*}[label=\roman*)]
    \item reduce the cost of labeling via using a small number of labeled samples for training, and
    \item query the class labels for the most informative subset of the data samples that are selected using an acquisition function.
\end{enumerate*} 
In other words, the best active learning strategy successfully selects certain unlabeled data samples from the distribution of available data, such that using this data portion for training leads to the maximal reduction of the classification error and variance. Thus, a wealth of work has been made on the development of various active learning approaches for different machine learning applications \cite{gal2017deep,hsu2015active}, and particularly for text classification \cite{mccallumzy1998employing,zhang2017active,zhu2008active} and toxicity detection \cite{breitfeller2019finding}. 

Meanwhile, the development of pre-trained language models, such as Google's BERT (Bidirectional Encoder Representations from Transformers) architecture \cite{devlin2018bert} and OpenAI's GPT-3 (Generative Pre-trained Transformer) model \cite{brown2020language}, which are trained over massive examples of written language have revolutionized the field of natural language understanding. Through knowledge sharing, these pre-trained models allow using a small amount of labeled data for fine-tuning the model for a downstream task \cite{dor2020active}, which makes these models ideal for active learning. Thus, the employment of such modern neural models with a practical active learning strategy seems to provide a promising direction via reducing the burden of manual labeling, and resulting in a strong performance across many diverse tasks.  

In this paper, we propose an active learning strategy for text classification. Specially, we are interested in mitigating the concerns associated with the annotation of online social media content with toxic language. To select the training samples in each data selection cycle, our methodology uses the learned feature space of the pre-trained BERT model to find the strongly coherent subsets of local clusters of data, known as the dominant sets. Then, the same number of training samples are drawn from the identified \emph{non-dominant} samples of each local cluster. These samples are more challenging to classify and represent the boundaries of the clusters in the embedding space. Thus, labeling non-dominant samples can provide the model with more information on the class boundaries. 
This is in contrast with many conventional uncertainty-based active learning methods. Such methods use the classification probabilities extracted from the model to calculate some measure of uncertainty and to select the most uncertain samples. However, as we will show since the model is trained on a few labeled data points, the uncertainty provided by the model itself is not a reliable metric. 

Based on our proposed methodology, we additionally propose a hybrid active learning strategy that allows us to incorporate the uncertainty score in the later stages of selection when the uncertainty score is more reliable. Our experiments using two different BERT-based text classification architectures and two datasets on toxic language classification suggest that the proposed selection strategies lead to higher performance. We also show that these strategies significantly outperform the other well-known active learning techniques, particularly in the early selection cycles and when the datasets are imbalanced.    
In summary, this paper makes the following contributions:
 \begin{itemize}
     \vspace{-2mm}   
     \item We propose a new metric of selection for the task of active learning using non-dominant sets (NDS).
     \item NDS has no parameters for fine-tuning, which makes it dataset- and problem-independent.
     \item We show how we can augment the NDS selection criteria to incorporate model uncertainty, whenever beneficial. 
     \item The superiority of the proposed algorithms is shown on different datasets and model architectures. 
 \end{itemize} 

\section{Background and Related Works}
\label{sec:background}
\subsection{Toxic Language in Online Social Media}
The problem of hate speech, and toxic language in general, is admittedly one of the major issues in online social media. Therefore, detecting and monitoring the propagation of such content has been a priority for social media companies. Studying the diffusion dynamics of content in online social media has revealed that hateful content spread much faster and farther than non-hateful content, reaching to significantly larger audience \cite{mathew2019spread}. This has motivated many social media companies to moderate such content to prevent its potentially disastrous consequences. 

It has also been shown that automated detection of hate speech is not a trivial task, as lexical detection methods cannot easily distinguish between hate speech and other instances of offensive language \cite{davidson2017automated}.  This signifies the need for high-quality datasets containing different instances of offensive language, as well as developing better methods of annotation and detection of cyberhate \cite{burnap2015cyber, waseem2016you}. For example, employing syntactic features has been shown to be useful to improve the identification of the targets and intensity of hate speech \cite{burnap2015cyber, gitari2015lexicon, silva2016analyzing, warner2012detecting}. However, in this work, our focus is on data collection and annotation. While collecting high-quality labeled datasets is crucial, annotating offensive and abusive content can be particularly discomforting for human annotators. In this work, we propose a model- and dataset-agnostic method for collecting the most informative instances of offensive language. We show that the number of annotated samples can be reduced significantly without reducing the classification performance.

\subsection{Active Learning}
Active learning refers to the process of efficient selection of the most informative data when the data is plentiful, but the labels are scarce \cite{settles2009active}. The use and the development of active learning methods for text classification are surging. For instance, \cite{kang2020active} proposed an active learning approach for a multi-label text emotion classification task using a probabilistic distance between the expected label distribution and uniform distribution. The main goal is to collect balanced data when the pool of the data is imbalanced. 

The task of named entity recognition from clinical text using active learning has been studied in \cite{wei2019cost}, which models the informativeness as well as the annotation cost. This approach was specifically designed for the scenarios where the labeling cost, e.g. time, for different samples is different. Therefore, an estimation of the labeling cost was also taken into the account for the selection method.

Zhang et al. \cite{zhang2017active} employed the embedding space of neural networks for word and sentence classification. The words or sentences that would potentially change the embedding function the most are selected to be labeled. This is estimated by calculating the expected gradient length. However, recent advancements in pre-training the embedding function with extremely large unlabeled datasets have eliminated the need for such consideration. For example, the application of active learning for binary text classification with the BERT pre-trained model has been investigated in \cite{dor2020active}, which provides prominent evidence that such pre-trained models are powerful tools for text classification when combined with active learning strategies. 

Here, we briefly review some of the most common acquisition functions in active learning literature. In this paper, we are interested in the pool-based active learning \cite{mccallumzy1998employing}, in which an acquisition function is used to query the label of a small set of selected samples. In this method, the model is initially trained using a small set of labeled data. Then, according to an informativeness criterion, the acquisition function selects a few data points to query their labels from the oracle. For a model $\mathit{\Mu}(x;\theta)$, pool data $\D_{pool}$, and input $x \in \D_{pool}$, the acquisition function $a(x,\Mu)$ is defined as:
\[
x^* = \arg\max_{x \in \D_{pool}} a(x, \Mu).
\]
The most informative samples are drawn from the pool and added to the training set by repeating the above step. For instance, investigating the uncertainty of the model as the informativeness criterion is a common approach for many active learning acquisition functions, with the hope that selecting samples based on such criterion leads to a lower model uncertainty. 

In this paper, we use various acquisition functions for comparison with the proposed strategy, which are reviewed below:

\begin{enumerate}[label=\Roman*.]
    \item Random Acquisition (baseline): This function selects data points uniformly at random from the pool of the unlabeled data. 
    
    \item Bayesian AL (Monte-Carlo Dropout):
    Proposed in \cite{gal2016dropout}, this method is an uncertainty-based AL strategy in which the class probability for each sample is approximated via calculating the average over $N$ inference cycles using Monte Carlo Dropout. 
    $$ \textit{Bayesian-AL} =  \frac{1}{N} \sum_n p(y=c | x, \theta_n) $$
    The authors in \cite{gal2016dropout} use Variation Ratios \cite{freeman1965elementary} as the acquisition function for their Bayesian AL strategy, defined as: 
    $$ \textit{VarRatio}(x) = 1 - \max_{y}\; p(y = c |x,\theta),$$
    in which larger values indicate a higher uncertainty score. Intuitively, the less probable our most probable class is, the more uncertain we are about the class of the sample.
    
    \item Minimum Margin: Proposed in \cite{scheffer2001active}, this approach is also an uncertainty-based acquisition function, which is more appropriate for a multiclass scenario. This function evaluates the uncertainty as the difference between the two most probable predictions as:
    $$\textit{MinMargin}(x) = p(y = c_1|x,\theta) - p(y = c_2|x, \theta),$$
    where classes $c_1$ and $c_2$ have the first and the second highest prediction scores. 
\end{enumerate}

\section{Methodology}
Our goal is to find the smallest set of training samples that lead to the maximal reduction of the classification error and variance. In this work, we are interested in the application of this strategy to text classification. Thus, we exploit the feature vectors from the BERT model to calculate the informativeness scores using our proposed acquisition functions described below. Since the BERT model is pre-trained over a huge amount of data points and can provide us with high-quality feature space, even before supervised fine-tuning, it is an ideal candidate for tasks such as active learning. 

\begin{figure*}
\centering
\includegraphics[width=0.9\textwidth]{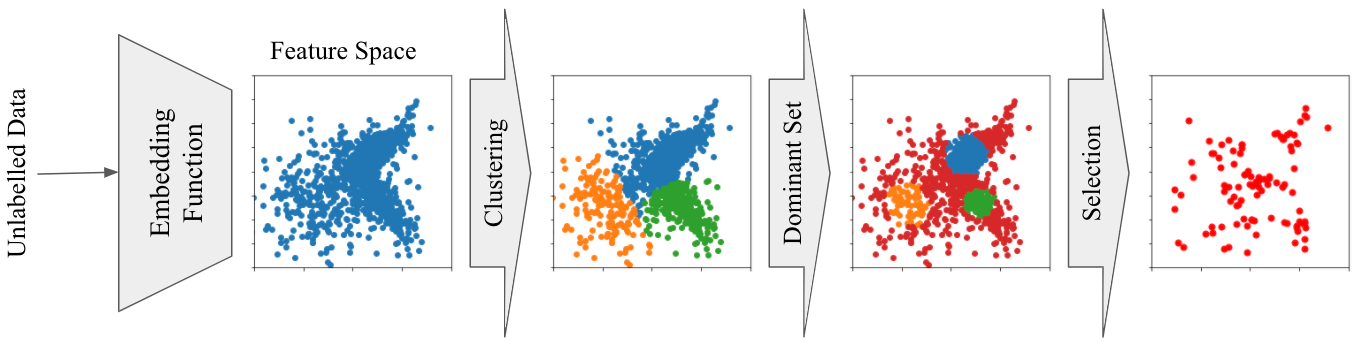}
\caption{\small Overview of the proposed NDS active learning framework for a toy $2$D feature space with $3$ classes. At each active learning cycle, an embedding function (e.g., pre-trained BERT) maps the input onto the feature space. Then, spectral clustering is used to cluster the extracted feature vectors. For each cluster, the dominant samples are identified by measuring the sample-cluster similarity. Finally, samples are selected randomly from the non-dominant set and labeled by an annotator. The selected samples represent the structure of the data well and most likely lie near the decision boundaries. 
}
\label{fig:overview}
\end{figure*}

The main intuition behind uncertainty-based methods is that if the model is uncertain about a sample, it likely lies near the decision boundary of the classes. Thus, knowing its label can help the classifier to better estimate the decision boundary. However, when the classifier is trained on only a few samples, or even no samples at all, the uncertainty score extracted from it is not reliable. In other words, the classifier does not yet know what it does not know. Therefore, we propose to exploit unsupervised methods such as clustering and dominant set detection to find these edge cases in the feature space of a pre-trained BERT model. For that, we first identify the local clusters of data in feature space and for each cluster detect the most similar set of points, referred to as the dominant set. The samples that do not belong to the dominant set of any of the clusters are the least similar samples to their corresponding clusters, and therefore the most challenging to classify. Furthermore, these samples are more likely to lie on the decision boundary. Thus, we can collect these informative samples without needing a classifier. Figure \ref{fig:overview} illustrates an overview of the steps of the proposed method for a toy $2$D feature space and $3$ classes. 
 
 \subsection{Active Learning Using Non-Dominant Set (NDS)}
 Here, we explain NDS in detail, which is a feature similarity-based active learning approach. Our selection strategy starts with spectral clustering of the feature vectors for the pool of unlabeled data to obtain a set of $K$ clusters $C_1 \dots C_K$ based on a pairwise local feature similarity score (e.g. Euclidean distance), where $K$ is equal to the number of classes in our classification problem. 
We extract a dominant set from each cluster $C_k$ by constructing an undirected edge-weighted graph $G_k(V_k, E_k, w_k)$ with no self-loops. $V_k$ and $E_k$ represent the set of vertices and edges per cluster $C_k$, and neighborhood relationships define the existence of edges. $a_{k}^{ij} = w_k(i,j)$ refers to the edge weight between feature vectors $i$ and $j$ if $(i, j) \in E_k$, and  $a_{k}^{ij} = 0$ otherwise. In other words, the vertices in the graph $G_k$ represent the samples in the $k^{\text{th}}$ cluster and the edges represent the distance between the samples. The proper distance function depends on the embedding function. In our experiments, where we employ BERT, Euclidean distance is used. 

To find the similarity of a sample $i$ to its corresponding cluster $C_k$, we can assign each vertex $i$ in the cluster a non-negative value $z_i$, representing the participation of the vertex in the cluster. The larger the $z_i$ the more corresponding node is associated with the cluster. If $z_i = 0$, then the sample $i$ is not associated with the cluster. This is a common way to represent the nodes in a cluster in classical graph-theoretical approaches \cite{shi2000normalized, sarkar1998quantitative,perona1998factorization} and a central quantity in dominant set identification \cite{pavan2006dominant}. The participation value $z_i, \forall i \in V_k$ can be found by solving the following optimization problem:
\begin{equation}
\label{eq:optimization}
\begin{aligned}
    \min \quad & z^T W_K z \\
    \text{subject to} \quad & z_i \geq 0 \\
    & \sum_i z_i = 1,
\end{aligned}
\end{equation}
where $z$ is a vector containing $z_1, z_2, \dots$ and $W_k$ is a matrix containing the pairwise distances $w_k(i,j)$.
Similar to \cite{pavan2006dominant},
we can use the replicator dynamics optimization technique, which is an evolutionary game theory approach, to extract $z_i$ associated with each node in the cluster. Using this optimization algorithm, the similarity of vertex $i$ with respect to the cluster $k$ is obtained. To divide the samples into dominant and non-dominant samples, we can use a threshold on the sample-cluster similarity $z_i$. For that, we use the median of the obtained positive similarity scores as the cutoff parameter, i.e., the median of the similarity scores after removing any zeros. This means that half of the points with non-zero participation are considered as the dominant set.  

Using this pairwise clustering and dominant set extraction method, it is expected that the identified dominant sets represent the highly compact structures within the embedding space, and thus, belong to the same class. We believe that using the embedding space of a pre-trained language model such as BERT is crucial to obtain clusters with a low amount of noise. With these assumptions, the non-dominant sets within each cluster $C_k$ contain more interesting samples for active learning as they hold less similarity to these maximally cohesive clusters. Accordingly, NDS concentrates on these sets from the pool data. To maximize the diversity over the embedding space and therefore the classes, we uniformly sample an equal number of data points from the non-dominant set of each cluster. Algorithm \ref{alg:NDS} shows the steps of the proposed NDS algorithm.

\begin{algorithm}
\caption{ Selection Using NDS}\label{alg:NDS}
\algsetup{
linenosize=\footnotesize,
linenodelimiter=:
}
\begin{algorithmic}[1]
\REQUIRE A pool of unlabeled data points $\D_{pool}$, number of samples to be selected $m$, number of classes $K$, an embedding function, and a distance function. \\
\hspace{-7mm} \textbf{Output}: Selected samples for annotation.\\
\STATE \textbf{Embedding:} Extract the feature vectors corresponding to all the points $x \in \D_{pool}$ using the embedding function. \\
\STATE \textbf{Clustering:} Extract the clusters $C_1, \dots, C_K$ using spectral clustering
\\ $ \text{for}\; k=1, \cdots ,K $
\STATE $\quad$ Calculate the pairwise distance matrix $W_k$
\STATE $\quad$ Calculate the sample-cluster similarity $z$ using \eqref{eq:optimization}
\STATE $\quad$ Calculate the threshold as $\tau = \textit{median}(z[z>0])$
\STATE $\quad$ Randomly select $\frac{m}{K}$ samples with $z_i \leq \tau$
\\ end 
\end{algorithmic}
\end{algorithm}

\subsection{Incorporating Uncertainty into NDS}
Although uncertainty-based methods such as variation ratio \cite{freeman1965elementary} and Bayesian AL \cite{gal2017deep} are widely used for active learning, these sampling strategies are known to have a higher tendency to the selection of outlier samples in the early cycles \cite{dor2020active}. As the size of training data increases in the later active learning cycles, the uncertainty-based methods are able to provide more reliable uncertainty scores.

On the contrary, we show that NDS is an effective method in the early sampling cycles as it selects the most critical samples from the pool. After a few active learning cycles, and specially, in the case of extremely imbalanced datasets, NDS can run out of the non-dominant set pool to select from. This is because the previously sampled training data is removed from the pool in every iteration, which results in the shrinkage of clusters. In such a scenario, we can increase the size of the non-dominant set by increasing the threshold of the dominant-set detection. 

We can also extend our approach via proposing NDS+, which is a compound sampling strategy that benefits from both NDS and uncertainty-based methods.
We can think of NDS as a random selection with equal weights over the non-dominant sets while setting the weight of dominant samples to $0$. Considering a uniform distribution over each of the identified per cluster non-dominant sets, the linear combination of NDS and uncertainty-based approaches becomes possible. We can define NDS+ with a smooth transition such that only NDS is used in the very early cycles, and the uncertainty scores influence the drawing procedure of samples in the later stages. Let $\Phi_U \in [ 0, 1 ]$ be an uncertainty measure, e.g. minimum margin, the hybrid sampling weight for NDS+ can be defined as:
\[ \alpha \Phi_{NDS} + (1-\alpha) \Phi_U, \]
where $\Phi_{NDS}$ is the NDS sampling weights as defined above and $\alpha$ is a parameter that regulates the relative effect of NDS selection versus the uncertainty-based strategy. The value of parameter $\alpha$ initially starts from $1$, the impact of uncertainty score can be gradually increased by reducing the parameter $\alpha$ over the cycles of active learning. 

\section{Experimental Setup}
\subsection{Dataset Description}
To evaluate the effectiveness of the proposed approach, we conduct our analysis using two datasets that we will briefly describe in this section. We refer the reader to the provided source citations for further details. 

The abusive language Twitter dataset  \cite{founta2018large} contains the \textit{abusive, hateful, spam}, and \textit{normal} classes. This dataset has been referred to as the Twitter-abusive dataset in our paper. The second dataset that we use in this paper is the \textit{Wikipedia Talk Labels: Personal Attacks}\cite{wulczyn2017wikipedia}, which belongs to the Wikipedia Detox Research Project. In this paper, we refer to this dataset as Wiki-attack. This dataset contains comments from Wikipedia talk pages, and is annotated for binary classification of \textit{attack} versus \textit{normal} classes. The key statistics for each dataset are provided in Table \ref{table:1}.

To prepare the model inputs for both datasets, first, we remove the emojis from the text corpus. Then, we replace the URLs with the special token ``HTTPURL". For the Twitter dataset, we also replace the usernames with the special token ``@USER". For the Wiki-attack text data, we remove the special tokens ``TAB\_TOKEN" and ``NEWLINE\_TOKEN". The data preparation procedure for the Wiki-attack dataset is the same as in \cite{wulczyn2017ex}. 
Next, we use the BERT tokenizer to convert the text inputs to sequences of tokens.
The train and test sets for the Twitter-abusive dataset are determined via randomly splitting this dataset with the ratio of 8:2. We used the predefined train and test splits for the Wiki-attack dataset.

\begin{table}[h!]
\centering
\begin{tabular}{| c || c | c |}
\hline
 Dataset & Class & Size \\ 
 \hline\hline 
&& \\ [-2ex]
 \multirow{6}{*}{\STAB{\rotatebox[origin=c]{90}{Twitter-abusive}}}
 && \\
  & Abusive & 22,766 \\ [0.5ex] 
 & Hateful & 4,496 \\  [0.5ex] 
  & Spam & 13,996    \\ [0.5ex] 
 & Normal & 53,560 \\ [2ex] 
 && \\ [-3ex]
 \hline\hline
%  && \\ [0.5ex]
   && \\ [-1ex]
 \multirow{3}{*}{\STAB{\rotatebox[origin=c]{90}{Wiki-attack}}}
 && \\
   & Attack & 13,542 \\ [0.5ex]
 & Normal & 10,1881 \\ [0.5ex] 
 && \\ [1ex]
\hline
\end{tabular}
\vspace{12pt}
\caption{The class distributions of the Twitter-abusive dataset (top) and the Wiki-attack dataset (bottom).}
\label{table:1}
\end{table}

\subsection{Model Configuration and Training Details}
To map the inputs to a feature space, we use the BERT-base architecture, which is known to capture the global as well as local context of text data. We use two different BERT-base architectures to conduct the experiments for text classification: \begin{enumerate*}[label=\roman*)]
\item the pre-trained BERT-base model, and
\item the pre-trained BERT-base model with 3 additional self-attention layers followed by a GRU layer. 
\end{enumerate*}
 The implementation details of the architecture (ii) follow the setting used in \cite{akula2021interpretable}. The additional self-attention layers for this architecture contain 8 heads, and the GRU layer outputs a 512-dimensional feature vector. We investigate architecture (ii) to examine the effect of extra attention layers and parameters for different active learning methods. A single dense layer as the classification head has been used for both architectures. 
 
Using the feature vectors from the {last hidden layer of the BERT-base model}, each text input $x$ is converted to a word embeddings $F \in \mathbb{R}^{S \times d}$, where $d = 768$ refers to the dimension of the embedding space, and $S$ is the maximal sequence length of the text inputs. We used $S = 50$, the learning rate of $2 \times 10^{-5}$, draw size of 20, and batch size of 64 across all the experiments. Hence, the same setting is used to train the two architectures using both datasets, exempting the number of training epochs, and initial training size. The models were trained for $e \in \{5, 10\}$ number of epochs. However, we observed a reduced performance in terms of F1 score and recall of all active learning methods for the architecture (i) when using 10 training epochs. This was specially noticed when the Wiki-attack dataset was used for training. On the contrary, higher performance scores were achieved for architecture (ii) using 10 training epochs. Thus, we report the experimental results using different epoch sizes for the two architectures.

  Similarly, we repeated the experiments using the initial training size of $n \in \{50, 100\}$ using the two datasets. When using the training size of 100 initial inputs from the Wiki-attack dataset, we observed that the performance of different active learning strategies in terms of F1 score represents a sharp jump in the early active learning cycles. As the comparison of different methods is more challenging under such a setting, we report the results for the initial training size of 50 and 100 for the Wiki-attack and Twitter-abusive datasets, respectively. The training sets of each dataset are used as the initial pool for data selection and label querying. For all the investigated active learning techniques, the first training iteration starts with an equal proportion of samples for each class.
  
\subsection{Active Learning Details}
After the initial training iteration, we use the model from the last training epoch for the evaluation with the test set, as well as the calculation of informativeness metrics by the acquisition functions. To evaluate the performance of the proposed approach, we perform comparative analysis using different active learning strategies,  details of which are provided in the background section. The minimum margin uncertainty scores are calculated via conducting one forward pass. However, the Bayesian AL uncertainty scores are obtained by averaging over 10 forward passes with an additional dropout unit with parameter 0.2, added before the fully-connected layer, as proposed in \cite{gal2016dropout}. Lastly, NDS and NDS+ use the learned embedding space of the BERT model to discover the non-dominant sets associated with each cluster. These two methods also use a cutoff parameter, which determines the dominant versus non-dominant sets of each cluster. This can become problematic when the size of identified non-dominant sets per associated class goes below $\frac{m}{k}$. We experienced this issue specially with extremely imbalanced data. Thus, we consider an adaptive cutoff value by increasing this parameter whenever the size of non-dominant set pools is less than $\frac{m}{k}$. In our experiments, we multiply the cutoff value by 10 until the size of the non-dominant set per cluster is sufficient for sampling. 
  
After the calculation of the informativeness scores for the pool of training samples, $m$ number of training inputs are drawn randomly from the pool of interest for each strategy. We use the minimum margin uncertainty scores for NDS+, and initially use $\alpha = 1$. This parameter is gradually decreased by 2\% at each active learning cycle. 

Next, the drawn samples are added to the training set, the models are reset, and the new set of training inputs are used to train a new model.   The explained procedure continues until a total of 500 training samples are selected and used for the training of the models. We conduct 10 runs for each experiment setting and report the average of the performance scores per active learning strategy. 

\begin{table*}
\footnotesize
\begin{center}
\begin{tabular}{|l|c|c|c|c|c|c|c|c|c|}
\hline 
Training Size & 140 & 160 & 180 & 300 & 340 & 400 & 500  \\
\hline\hline
Random   & $0.628 $ & $0.609 $ & $0.527 $ & $0.704 $ & $0.711 $ & $0.735 $ & $0.752 $ \\
Minimum Margin   & $0.582 $ & $0.524 $ & $0.554 $ & $0.703 $ & $0.719 $ & $0.748 $ & $0.766 $ \\
Bayesian AL   & $0.540 $ & $0.504 $ & $0.470 $ & $0.694 $ & $0.717 $ & $0.745 $ & $0.761 $ \\
NDS   & $\mathbf{0.666 }$ & $\mathbf{0.667}$ & $\mathbf{0.628 }$ & $\mathbf{0.740 }$ & $\mathbf{0.757 }$ & $0.749 $ & $0.754 $ \\
NDS+   & $0.664 $ & $0.660 $ & $0.607 $ & $0.723 $ & $0.755 $ & $\mathbf{0.762 }$ & $\mathbf{0.775 }$ \\
\hline
\end{tabular}
\end{center}
% \vspace{-5mm}
\caption{\small Average of classification F1 scores for the Twitter-abusive dataset using architecture (i) with 10 training epochs. Early (training size of 140, 160, 180), middle (training size of 300, 340), and later (training size of 400, 500) active learning cycles are reported. All methods use the same initial training data of size 100 samples with equal class distributions. 
}
\label{table:2}
% \vspace{-5mm}
\end{table*}

\begin{table*}
\footnotesize
\begin{center}
\begin{tabular}{|l|c|c|c|c|c|c|c|c|c|}
\hline 
Training Size & 140 & 160 & 180 & 300 & 340 & 400 & 500  \\
\hline\hline
Random   & $0.634 $ & $0.679 $ & $0.681 $ & $0.729 $ & $0.727 $ & $0.730 $ & $0.758 $ \\
Minimum Margin   & $0.622 $ & $0.659 $ & $0.686 $ & $0.734 $ & $0.738 $ & $0.755 $ & $0.772 $ \\
Bayesian AL   & $0.604 $ & $0.653 $ & $0.641 $ & $0.734 $ & $0.736 $ & $0.749 $ & $0.763 $ \\
NDS   & $0.651 $ & $\mathbf{0.683}$ & $0.683 $ & $0.749 $ & $0.758 $ & $0.766 $ & $0.765 $ \\
NDS+   & $\mathbf{0.659} $ & $0.681 $ & $\mathbf{0.688} $ & $\mathbf{0.750} $ & $\mathbf{0.763} $ & $\mathbf{0.772 }$ & $\mathbf{0.779 }$ \\
\hline
\end{tabular}
\end{center}
% \vspace{-5mm}
\caption{\small Average of classification F1 scores for the Twitter-abusive dataset using architecture (ii) with 10 training epochs. Early (training size of 140, 160, 180), middle (training size of 300, 340), and later (training size of 400, 500) active learning cycles are reported. All methods use the same initial training data of size 100 samples with equal class distributions. 
}
\label{table:3}
% \vspace{-5mm}
\end{table*}

\section{Experimental Results}
In this section, we report the results of our proposed active learning methods, NDS and NDS+, in comparison to random selection, minimum margin \cite{scheffer2001active}, and Bayesian AL with variation ratio acquisition function \cite{gal2017deep}. 

Tables number \ref{table:2} and \ref{table:3} report the classification F1 scores for the explored active learning strategies using architectures (i) and (ii). The reported scores are the average of the classification F1 scores over 10 training runs using $n \in \{140, 160, 180, 300, 340, 400, 500\}$ training samples selected from the \textit{abusive language Twitter dataset} \cite{founta2018large}. We particularly report on these training sizes to cover a comparison of performance scores for early (training size of 140, 160, 180), middle (training size of 300, 340), and later (training size of 400, 500) active learning cycles. From these tables, it is evident that the two sampling strategies that concentrate on the dominant sets for data selection outperform the other methods. 

\begin{table*}[!ht]
\footnotesize
\begin{center}
\begin{tabular}{|l|c|c|c|c|c|c|c|c|c|}
\hline 
Training Size & 90 & 130 & 170 & 290 & 350 & 450 & 500  \\
\hline\hline
Random   & $0.734 $ & $0.709 $ & $0.712 $ & $0.741 $ & $0.750 $ & $0.752 $ & $0.737$\\
Minimum Margin   & $0.688 $ & $0.701 $ & $0.596 $ & $0.703 $ & $0.758 $ & $0.766 $ & $0.745 $ \\
Bayesian AL   & $0.596 $ & $0.595 $ & $0.625 $ & $0.704 $ & $0.719 $ & $0.747 $ & $0.734 $ \\
NDS   & $0.736 $ & $\mathbf{0.727}$ & $0.731 $ & $\mathbf{0.770 }$ & $\mathbf{0.769 }$ & $0.759 $ & $0.752 $ \\
NDS+   & $\mathbf{0.738} $ & $0.741 $ & $\mathbf{0.737} $ & $0.723 $ & $\mathbf{0.769} $ & $\mathbf{0.777 }$ & $\mathbf{0.783 }$ \\
\hline
\end{tabular}
\end{center}
% \vspace{-5mm}
\caption{\small Average of classification F1 scores for the Wiki-attack dataset using architecture (i) with 5 training epochs. Early (training size of 90, 130, 170), middle (training size of 290, 350), and later (training size of 450, 500) active learning cycles are reported. All methods use the same initial training data of size 50 samples with equal class distributions. 
}
\label{table:4}
% \vspace{-5mm}
\end{table*}

\begin{table*}[!ht]
\footnotesize
\begin{center}
\begin{tabular}{|l|c|c|c|c|c|c|c|c|c|}
\hline 
Training Size & 90 & 130 & 170 & 290 & 350 & 450 & 500  \\
\hline\hline
Random          & $0.781 $ & $0.776 $ & $0.752 $ & $0.755 $ & $0.761 $ & $0.774 $ & $0.780$\\
Minimum Margin  & $0.771 $ & $0.760 $ & $0.664 $ & $0.716 $ & $0.759 $ & $0.786 $ & $0.795 $ \\
Bayesian AL     & $0.744 $ & $0.686 $ & $0.648 $ & $0.724 $ & $0.729 $ & $0.751 $ & $0.787 $ \\
NDS   & $\mathbf{0.802} $ & $\mathbf{0.790}$ & $0.779 $ & $0.792 $ & $\mathbf{0.825 }$ & $0.799 $ & $0.806 $ \\
NDS+   & $0.792$ & $0.777 $ & $\mathbf{0.785} $ & $\mathbf{0.801 }$ & $0.804 $ & $\mathbf{0.822 }$ & $\mathbf{0.819 }$ \\
\hline
\end{tabular}
\end{center}
% \vspace{-5mm}
\caption{\small Average of classification F1 scores for the Wiki-attack dataset using architecture (ii) with 5 training epochs. Early (training size of 90, 130, 170), middle (training size of 290, 350), and later (training size of 450, 500) active learning cycles are reported. All methods use the same initial training data of size 50 samples with equal class distributions. 
}
\label{table:5}
% \vspace{-5mm}
\end{table*}

Similarly, the results achieved for architectures (i) and (ii) using the \text{Wikipedia talk labels: personal attacks dataset} \cite{wulczyn2017wikipedia} confirm the superiority of the dominant-set based methods over other active learning strategies. The classification F1 scores of the two architectures using different active learning strategies trained over this dataset are available in Tables \ref{table:4} and \ref{table:5}. These scores are averaged over 10 runs, and using training size of $n \in \{90, 130, 170, 290, 350, 450, 500 \}$. The performance scores that are reported refer to 5 and 10 training epochs for the Wiki-attack and Twitter-abusive datasets, respectively. 

From the result tables, it can be observed that the uncertainty-based active learning methods, Bayesian AL and minimum margin, fall behind the other selection strategies in the first few active learning cycles. In fact, it is evident that the random acquisition function leads to higher classification F1 scores for both architectures in the early stages of data sampling. This is consistent with the results and analysis of uncertainty-based methods in the literature of active learning, as these sampling strategies are known to have a higher tendency to the selection of outlier samples in the early cycles \cite{dor2020active}. 
However, as the size of training data grows, the uncertainty-based methods seem to manage to produce more reliable uncertainty scores, and thus, discover more critical samples for training. 

On the contrary, the two non-dominant set-based methods, NDS and NDS+, are able to select the critical samples for training, even in the early cycles of active learning. This difference is more significant when comparing different active learning methods to fine-tune the BERT-base model, as shown for architecture (i). On the other hand, the three additional self-attention layers, as well as the GRU layer in architecture (ii) seem to manage to compensate for the deficiency of the selection strategies, with the cost of higher computational complexity. 
Further comparison of the obtained results for architecture (i) versus (ii) also suggests that the additional layers in architecture (ii) result in having a model that is more robust to noise and less sensitive to outlier samples. This is specially noticeable for uncertainty-based sampling strategies.

Another important factor to discuss about the non-dominant set-based strategies is that the sizes of non-dominant sets shrink over the cycles of active learning with this technique, resulting in a smaller pool of samples for selection in the next iterations. This is specially a concern when using an extremely imbalanced dataset. Accordingly, the compound strategy that exploits an independent score, such as model uncertainty scores, must achieve superior results over the long run. This is noticeable from the obtained results for NDS+ in Tables \ref{table:2}-\ref{table:5}. Specially, our results represent the superiority of this method for later active learning cycles as expected. However, the last 3 tables also suggest that this method outperforms other strategies in early and middle active learning cycles, possibly, as a result of considering a more diverse pool of samples for selection. 

Comparing the results for Table \ref{table:2} versus \ref{table:4}, which belong to the classification F1 scores obtained by architecture (i) suggest that NDS+ is indeed taking advantage from both NDS and uncertainty-based selection methods. As this technique represents a competing performance with NDS in Table \ref{table:3}, we believe that the superiority of NDS in the early and middle active learning cycles for this specific experiment can be replaced by NDS+ if we set the importance parameter $\alpha$ differently. For instance, starting to decrease the weight of NDS in a later active learning cycle, such that our selected samples rely on the non-dominant sets for a longer time. Additionally, acknowledging the sharp decline in the F1 scores associated with the uncertainty-based strategies in the early active learning cycles, the significance of using non-dominant sets for the selection pool becomes even more evident. 

Our extensive analysis and comparison of the proposed methods with other active learning approaches using two different text classification corpus, and two deep architectures suggest that randomly selecting from the non-dominant set associated with each cluster is a powerful strategy to reduce the labeling cost and effort. That being said, we should that further improvements can be achieved via considering a joint strategy, as reported for NDS+.

\section{Conclusions}
In this work, we presented a new criterion to select the most informative samples from a pool of unlabeled data points. Our proposed algorithm detects the local clusters in the embedding space and selects the samples that are not strongly coherent with the cluster, i.e., the non-dominant set. Our numerical results illustrate the effectiveness of this selection technique for the task of hate speech detection in online social media on different datasets and classification models. Specifically, we show that our method outperforms the state-of-the-art uncertainty-based methods in the early stages of selection when the uncertainty score extracted from the model is not accurate. We also propose a hybrid algorithm that is able to incorporate the uncertainty score into its decision criteria in the later stages of selection. Our results and analysis suggest that an active learning task can be divided into two distinct phases. In the early stages, unsupervised techniques such as employing pre-trained models, clustering, and identifying the dominant sets outperform the supervised methods, i.e., uncertainty score extracted from the trained model. However, in the second phase, later stages of selection, taking the model uncertainty into the account can improve the selection performance. Such hybrid methods are not currently well-studied and we believe that our encouraging results will stimulate further research on similar active learning strategies.

\section{Ethics Statement}
Our proposed active learning approach has the potential to mitigate the difficulties associated with the annotation, and classification of textual content, e.g. annotation cost and bias. The task of labeling offensive and abusive content is particularly difficult, as it can cause discomfort and emotional disturbance in the human annotators. Via extensive experiments and analysis, we show that our method is particularly practical in the classification of toxic language in online social media using a small amount of labeled data. However, our method can be used in other fields as well to facilitate and accelerate the development of AI research.  The experiments were designed with an effort to be fair. To the best of our knowledge, our method does not have any specific drawbacks or limitations.

\bibliographystyle{unsrt}
\bibliography{references.bib}

\end{document}